\documentclass[11pt]{article}

\usepackage[final]{acl}

\usepackage{times}
\usepackage{latexsym}

\usepackage[T1]{fontenc}

\usepackage[utf8]{inputenc}

\usepackage{microtype}

\usepackage{inconsolata}

\usepackage{graphicx}

\usepackage{booktabs} 
\usepackage{amsmath}  
\usepackage{graphicx} 
\usepackage{multirow} 
\usepackage{tabularx}

\usepackage{enumitem}

\usepackage{float}

\title{The ``Label Effect'': Source Bias and Alignment in Trust Assessment by Human and LLM-as-a-Judge}

\title{Label Effect: Human-Aligned Heuristic Biases in Trust Assessment by LLM-as-a-Judge}

\title{Label Effects: Shared Heuristic Reliance in Trust Assessment by Humans and LLM-as-a-Judge}

\author{
     \textbf{Xin Sun\textsuperscript{1,2}},
     \textbf{Di Wu\textsuperscript{2}},
     \textbf{Sijing Qin\textsuperscript{1,4}},
     \textbf{Isao Echizen\textsuperscript{1,3}},
     \textbf{Abdallah El Ali\textsuperscript{5,6}},
     \textbf{Saku Sugawara\textsuperscript{1,3}}
     \medskip
     \\ 
     \textsuperscript{1}National Institute of Informatics (NII), Japan\\
     \textsuperscript{2}University of Amsterdam, the Netherlands\\
     \textsuperscript{3}University of Tokyo, Japan\\
     \textsuperscript{4}Hitotsubashi University, Japan\\
     \textsuperscript{5}Centrum Wiskunde \& Informatica (CWI), the Netherlands\\
     \textsuperscript{6}Utrecht University, the Netherlands\\
}

\begin{document}
\maketitle
\begin{abstract}
Large language models (LLMs) are increasingly used as automated evaluators (LLM-as-a-Judge). 
This work challenges its reliability by showing that trust judgments by LLMs are biased by disclosed source labels.
Using a counterfactual design, we find that both humans and LLM judges assign higher trust to information labeled as human-authored than to the same content labeled as AI-generated.  
Eye-tracking data reveal that humans rely heavily on source labels as heuristic cues for judgments. 
We analyze LLM internal states during judgment. Across label conditions, models allocate denser attention to the label region than the content region, and this label dominance is stronger under Human labels than AI labels, consistent with the human gaze patterns. 
Besides, decision uncertainty measured by logits is higher under AI labels than Human labels. 
These results indicate that the source label is a salient heuristic cue for both humans and LLMs. 
It raises validity concerns for label-sensitive LLM-as-a-Judge evaluation, and we cautiously raise that aligning models with human preferences may propagate human heuristic reliance into models, motivating debiased evaluation and alignment.
\end{abstract}

\section{Introduction}

\begin{figure*}[!ht]
    \centering
    \vspace{-1.2mm}
    \includegraphics[width=0.96\textwidth]{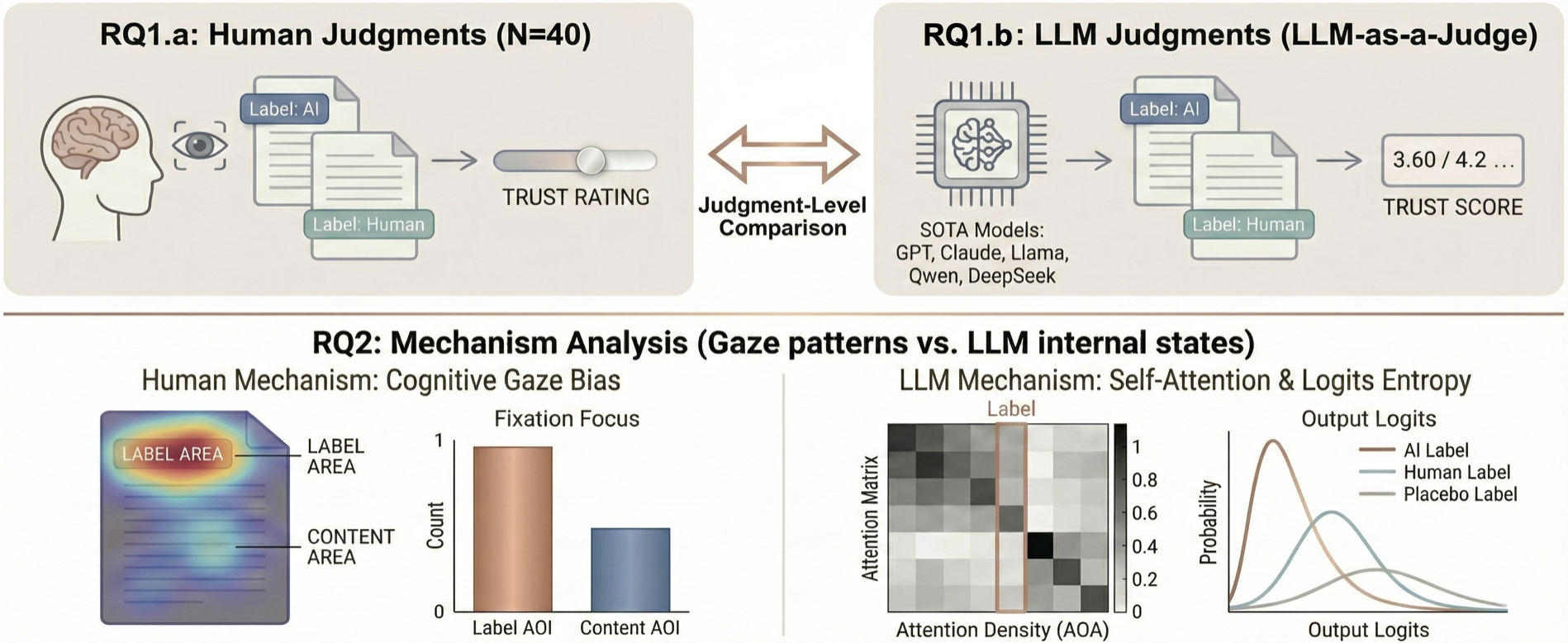} 
    \caption{Three studies examine how source labels affect trust judgments by an LLM-as-a-judge and investigate the alignment with human behavior via mechanistic analyses of gaze attention, model self-attention, and logits entropy.}
    \label{fig:study_procedure}
    \vspace{-2.6mm}
\end{figure*}


In the era of large language models (LLMs), health information is easier to access. 
However, non-experts often struggle to judge trustworthiness as it requires expertise. 
Thus, they frequently show reliance on heuristic cues such as source credibility (e.g., ``Written by an Expert'')~\cite{journalmedia5020046} for trust assessment. 
Such labels are double-edged: they can foster trust~\cite{scharowski_certification_2023} but also undermine it~\cite{ai_people_heard_label,ai_profile_text}, showing that trust judgments are sensitive to these heuristic cues~\cite{responsible_ai}.
We refer to this phenomenon as the \emph{``Label Effect''}.
While heuristics reduce cognitive effort, they also bias evaluation: strong labels may overshadow content and enable ``blind trust", leading to the acceptance of low-quality advice based on perceived authority or compliance-related AI disclosures (e.g., EU AI Act)~\cite{Elali2024}. This risk is amplified in health and generative-AI settings, where provenance can be mismatched (e.g., AI-generated advice presented as expert-written), leading to real-world harm. 

While LLMs are increasingly used as a judge for scalable automated evaluations~\cite{li2024llms,li-etal-2025-generation}, with the assumption that LLM judgments can objectively reflect quality of the evaluated content~\cite{gu2025surveyllmasajudge},
many studies argue that LLM-as-a-Judge evaluations exhibit bias in practice~\cite{ye2024justiceprejudicequantifyingbiases}, including inconsistency~\cite{wang2025trustjudgeinconsistenciesllmasajudgealleviate,haldar-hockenmaier-2025-rating}, limited reliability~\cite{schroeder2025trustllmjudgmentsreliability}, self-preference~\cite{wataoka2025selfpreferencebiasllmasajudge}, reasoning bias~\cite{wang2025assessingjudgingbiaslarge} and scoring bias~\cite{li2025evaluatingscoringbiasllmasajudge}.
For high-stakes health information, it remains unclear whether source labels bias trust scoring by LLM-as-a-Judge and how these label-driven shifts are consistent with human heuristic-driven trust patterns~\cite {responsible_ai}. 
This motivates our first research question:
\begin{enumerate}[label=, leftmargin=*, itemsep=0pt]
    \vspace{-1.2mm}
    \item \textit{\textbf{(RQ1: judgment-level)} In high-stakes health contexts, to what extent do disclosed source labels affect trust judgments by (a) humans and (b) LLM-as-a-Judge under counterfactual label swaps for identical health information?}
    \vspace{-1.6mm}
\end{enumerate}

To ground our investigation, we first conduct a controlled human study with a counterfactual design that isolates label effects. 
Participants rate their trust in identical health information presented with different source labels (i.e., as either ``human-authored'' or ``AI-generated''). 
We found that Human-labeled information consistently received higher trust than AI-labeled information. 
LLM-as-a-Judge exhibits the same judgment-level label effect: LLMs assign higher trust to Human-labeled information than to AI-labeled information.

These findings raise important practical concerns.
Alignment techniques aim to shape model behavior toward human preferences~\citep{ouyang2022training,jiang2024survey}, assuming that aligning model behaviors with human values yields safer, more rational systems. 
However, this assumption overlooks that human cognition itself may contain heuristic reliance or bias.
If humans prioritize trust heuristic cues like source labels for judgment~\cite{Reis_Moritz}, then aligning models to human preferences may risk encoding this bias directly into models' generation and reasoning capabilities, making LLM judges vulnerable to label spoofing and less reliable as impartial evaluators. 
We then go beyond judgment-level behaviors and keep asking:
\begin{enumerate}[label=, leftmargin=*, itemsep=0pt]
    \vspace{-1.4mm}

    \item \textit{\textbf{(RQ2: mechanism-level)} How do source labels alter internal processing signals during trust assessment in humans (gaze) and LLM-as-a-Judge (attention and uncertainty)?}
\vspace{-1.4mm}
\end{enumerate}
We explore how label effects manifest at the processing level by analyzing (i) human gaze patterns across AoIs (Areas of Interest) and labels; and (ii) LLM's attention and inference uncertainty~\cite{sheng-etal-2025-analyzing}, while noting that these signals provide correlates rather than causal explanations for both humans~\cite{Cacioppo} and LLM-as-a-Judge~\cite{wiegreffe-pinter-2019-attention}.


From our analysis, LLM-as-a-Judge exhibits a similar label-driven pattern during trust assessment following the same directional reliance observed in human ratings. 
Specifically, LLMs allocate denser attention to the label region than the content region, and this label dominance is stronger for the Human label than the AI label.
This is consistent with the human gaze patterns that participants fixate more on the label area under Human labels than AI labels, suggesting that label semantics serve as a heuristic cue for both humans and LLM judges. 
It raises a broader concern that human preference-based alignment may inherit such human shortcut cues~\cite{Brady2025-fi,llms_amplified_cognitive_biases}, making LLM judges vulnerable to label spoofing.

Our work makes three contributions. 
(1) We provide controlled empirical evidence that source labels bias trust judgments in both human and LLM-as-a-Judge evaluators, raising validity concerns. 
(2) We investigate internal patterns of LLMs associated with label effects. 
(3) By grounding LLM judgments with humans, we show that aligning human preference may propagate human biases into LLMs.
These results have direct implications for the design of LLM-as-a-Judge evaluation and for model alignment with human values. 
We provide materials and codes for reproducibility\footnote{\url{https://anonymous.4open.science/r/Label-Effects/}}.


\section{Related Work}

\subsection{Source Credibility Cues for Human Trust}

Trustworthiness assessment is cognitively demanding, especially for non-experts in high-stakes domains such as healthcare. 
A large body of psychology and communication research shows that when readers cannot easily verify accuracy, they rely on \emph{heuristic cues}, such as source credibility signals (e.g., authority, expertise disclosures), as shortcuts for trust judgments~\cite{responsible_ai}. 
Dual-process theory~\cite{dual_process} echoes that such cues can dominate when ability to scrutinize content is limited. 
Empirically, source labels and credibility badges can shift perceived trust even when content is unchanged, increasing risks to misleading or spoofed attribution~\cite{ai_people_heard_label}.

Despite these insights, it remains unclear how these cue-driven trust judgments relate to emerging LLM-based judgments. 
We address this gap by grounding label effects in a controlled eye-tracking study, which we compare against LLM-as-a-Judge behaviors under identical label manipulations.


\subsection{LLM-as-a-Judge and Evaluation Biases}
Using LLMs as automatic evaluators (i.e., LLM-as-a-Judge) has become a practice for scalable evaluation~\cite{li2024llms,gu2025surveyllmasajudge}. 
Prior work also shows that LLM judges can be fragile: ratings may shift with prompt framing~\cite{li2025llmsreliablyjudgeyet}, positional structures~\cite{shi2025judgingjudgessystematicstudy}, or self-preferences~\cite{spiliopoulou2025playfavoritesstatisticalmethod,llm_prefer_ai_generated,retrievers_prefer_ai_generated}, factors unrelated to the target quality signal.
These findings indicate that LLM-as-a-Judge scores are not purely objective, but can be shaped by evaluation artifacts.

Recent work~\cite{marioriyad2025silentjudgeunacknowledgedshortcut,chen-etal-2024-humans} shows that authority cues can bias LLM-as-a-Judge. 
However, the effects of true source and labels (Human vs.\ AI) remain unclear in high-stakes health trust assessment, and most prior work examines only output scores without probing internals or contrasting LLM behavior with human behaviors. 
We investigate deeper to address these gaps.



\subsection{Alignment of LLMs with Human Values}

Current LLM alignment methods aim to make model behavior consistent with human and safety expectations by incorporating human-derived preferences. 
A dominant approach is Reinforcement Learning from Human Feedback (RLHF)~\cite{ouyang2022training}, which uses human preferences to instruct models (e.g., InstructGPT). 
Several studies report agreement between LLM-as-a-Judge and humans~\cite{liu2023gevalnlgevaluationusing,zheng2023judgingllmasajudgemtbenchchatbot}. 
While others argue that LLMs fail to reflect human-like behavior~\cite{tjuatja-etal-2024-llms}.

Recent work indicates that LLMs can produce both heuristic, bias-prone responses as humans. 
However, these behaviors are not equivalent to human cognition, as ``cognitive'' biases in LLMs likely reflect reward bias from training data~\cite{Brady2025-fi}.
We connect it to provenance cues in this work: if humans rely on source labels as shortcuts, aligning models to human preferences may amplify the same reliance in LLM-as-a-Judge.


\begin{figure*}[!ht]
    \centering
    \includegraphics[width=0.96\textwidth]{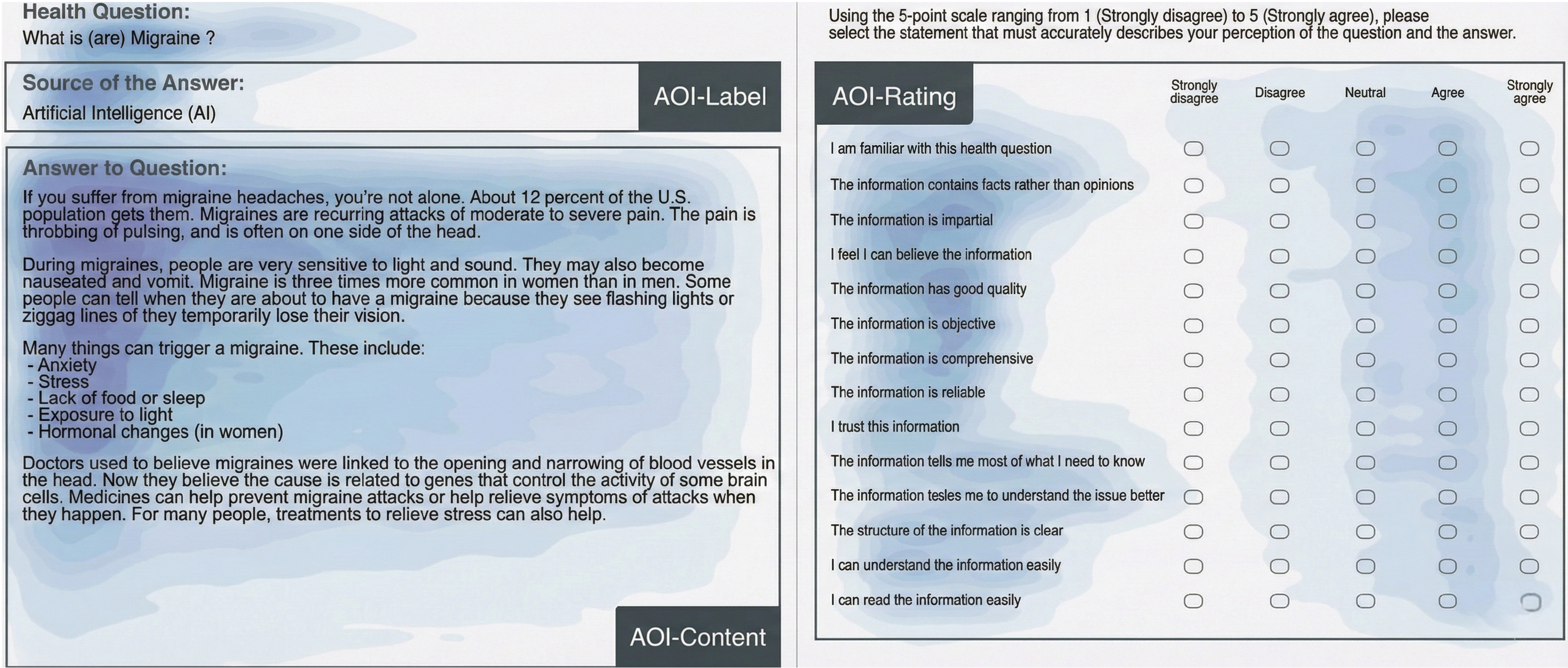} 
    \vspace{-1mm}
    \caption{
    An example heatmap of gaze points on stimuli that are displayed on the lab monitor during the human assessment. 
    Three AoIs are predefined: AoI-Label is the area for presenting disclosed label; AoI-Content is the area for presenting health information; AoI-Rating is the area to rate trust level in information on the left side.
    }
    \label{fig:AoI_setup}
    \vspace{-3mm}
\end{figure*}


\section{Methodology}
\label{sec:methodology}

\paragraph{Experimental Design and Dataset}
\label{subsec:experimental_design}

We adopt 150 health question-answer (QA) pairs from dataset~\cite{CHQ_Summ}.
Each data point is denoted as a tuple $D = (Q, A, L)$, where $Q$ represents the health question, $A$ is the answer to the question, and $L$ is the source label of the answer. 

To separate the effects of \textit{true answer source} from the effects of \textit{disclosed source labels}, we use a controlled $2\times2$ factorial design with counterfactual variants $(Q, A, L)$ that differ in both $A$ and $L$.
Because the \emph{same answer text} is presented under different labels (congruent or incongruent with its source), this design enables counterfactual comparisons that test the \emph{label effect} while holding content constant.
Below are the manipulations we did: 

\begin{itemize}[leftmargin=*, itemsep=2.6pt, topsep=2.6pt, parsep=0pt]
    \item \textbf{Source of Content ($S$) (i.e., actual source):}
    \begin{itemize}[leftmargin=*, itemsep=0pt, topsep=0pt, parsep=0pt]
        \item $S_{human}$: the answer was written by Human Professionals from the selected dataset.
        \item $S_{LLM}$: the answer was generated by GPT-4o~\cite{gpt4} with length/format constraints to match expert-authored answers.
        To reduce confounds, domain experts reviewed the GPT-generated answers to ensure they were comparable to $S_{human}$ in relevance to the length, presentation format and relevance.
    \end{itemize}

    \item \textbf{Label of Source ($L$) (i.e., manipulated source):} 
    \begin{itemize}[leftmargin=*, itemsep=0pt, topsep=0pt, parsep=0pt]
        \item $L_{human}$: A human label (``Human Authored'').
        \item $L_{AI}$: An AI label (``AI-Generated'').
    \end{itemize}
\end{itemize}

\paragraph{Tasks and Conditions}
\label{subsec:experimental_conditions}
Human participants and LLM-as-a-Judge models perform the same trust judgment task. 
Given a tuple $(Q, A, L)$, evaluators rate their perceived trust in both information on a 5-point Likert scale, adapted from validated \textit{Trust of Online Health Information} questionnaire~\cite{ti,Rowley2015StudentsTJ}, with multiple trust-related items (e.g., credibility, reliability) as shown in Fig.~\ref{fig:AoI_setup}.
We aggregate and average the ratings of these items as the trust score.



\paragraph{Study Overview}
We conduct the following studies (see Fig.~\ref{fig:study_procedure}) that map directly to our RQs:
\vspace{-1.2mm}
\begin{itemize}[leftmargin=*, itemsep=2.6pt, topsep=2.6pt, parsep=0pt]
    \item \textbf{Human grounding: label effects and gaze attention} (Sec.~\ref{subsec:human_study}).
    We quantify how disclosed labels influence human trust ratings (RQ1.a) and gaze between Label and Content AoIs (Area of Interest), establishing a cognitive behavioral reference of reliance on trust heuristic cues.

    \item \textbf{LLM-as-a-Judge: judgment-level label effects} (Sec.~\ref{subsec:llm_study}).
    Using counterfactual label swaps with identical content, we test whether LLM trust ratings differ across labels (RQ1.b).

    \item \textbf{Mechanism analysis of label effects}
    (Sec.~\ref{subsec:mechanistic_analysis})
    We analyze LLM attention allocation between Label and Content regions and decision uncertainty via logits, enabling human-LLM alignment comparisons (RQ2).
    We further validate the semantic label effects using a placebo label condition.
\end{itemize}


\subsection{Human Grounding of Label Effects with Eye-Tracking}
\label{subsec:human_study}

We conducted a laboratory study employing eye-tracking to provide human grounding by examining whether source labels affect trust judgment (\textit{RQ1.a}) and how gaze attention is allocated during human cognitive processing (\textit{RQ2}).

\paragraph{Participants and Procedure}
We recruited participants ($N=40$) with non-medical backgrounds (demographics in Appendix~\ref{sec:appendix_lab_demographics}).
Each participant read 12 QA pairs from two actual sources (\emph{$S_{human}$ vs.\ $S_{LLM}$}) manipulated by two labels (\emph{$L_{human}$ vs.\ $L_{AI}$}) in counterbalanced order (Sec.~\ref{subsec:experimental_design}). 
Participants read each QA pair at their own pace and then rated trust.
To avoid memory effects, no participant saw the same answer under both labels.
Gazes were recorded with a screen-based eye-tracker (sampling rate: 60 Hz) (setups in Appendix~\ref{sec:appendix_lab_setups}).

\paragraph{Areas of Interest (AoIs) for Gaze Attention}
To quantify attention allocation during judgment, we defined two non-overlapping AoIs for each $(Q, A, L)$ as shown in Fig.~\ref{fig:AoI_setup}:

\begin{itemize}[leftmargin=*, itemsep=2.6pt, topsep=2.6pt, parsep=0pt]
    \item \textbf{Label AoI ($AoI_{\mathrm{label}}$):} The region displaying the disclosed source label (e.g., “Human-authored” or “LLM-generated”).
    
    \item \textbf{Content AoI ($AoI_{\mathrm{content}}$):} The region containing the answer text to the health question.
\end{itemize}



As gaze metrics, we use \emph{Fixation Count} and \emph{Fixation Duration}~\cite{fixations} as proxies for visual attention, defined as the number of fixations and total dwell time within each AoI.


\subsection{Judgment-Level Label Effect of LLMs}
\label{subsec:llm_study}

This study specifically tests \textit{RQ1.b}: whether source labels affect trust ratings by LLMs. 
We applied the same design used in the human experiments.

\paragraph{Models}
We evaluate a diverse set of proprietary and open-sourced LLMs (detailed in Appendix~\ref{sec:appendix_llms_details}):
\begin{itemize}[leftmargin=*, itemsep=2.6pt, topsep=2.6pt, parsep=0pt]
    \item \textbf{Proprietary LLMs}: GPT-5.2, GPT-4o, Claude-Sonnet-4.5, Claude-Opus-4.5.
    \item \textbf{Open-weight LLMs}: GPT-OSS-120B, LLaMA-3.2-70B, Qwen3-235B, DeepSeek-v3.2-685B.
\end{itemize}


\paragraph{Prompt Design}
We designed the prompt 
to closely mirror the instructions given to human evaluators. 
Each model receives a tuple $(Q, A, L)$ consisting of a health question $Q$, an answer $A$, and a source label $L$.
The label $L$ is manipulated, while $Q$ and $A$ are identical across label conditions.
Models were instructed to rate the trust scores on a 5-point scale using the same dimensions as in human experiments (Fig.~\ref{fig:AoI_setup}), enabling direct comparison between human and LLM judgments. 
We set the decoding temperature to 0 to minimize sampling randomness and make condition differences attributable to label manipulations.
Full prompt templates are provided in Appendix~\ref{sec:appendix_prompt}.






\subsection{Mechanism-Level Analysis of Label Effect}
\label{subsec:mechanistic_analysis}

This study addresses \textit{RQ2} by probing how disclosed source labels influence LLMs' judgment internally, and by enabling AoI-level comparisons to human gaze patterns. 
We analyze two LLM internal states: 
(i) \textbf{attention allocation} between Label and Content regions, and 
(ii) \textbf{judgment uncertainty} quantified by logits.
We test these \textbf{open-weighted models}: \emph{LLaMA-3.2-3B \& 70B, GPT-OSS-20B \& 120B, and Qwen3-30B normal \& instructed versions} (Details in Appendix~\ref{sec:appendix_llms_details}).


\subsubsection{Attention Allocation}
\label{attention_analysis}

\paragraph{AoA-level attention density with same label}
For each model and label (Human, AI), we extract last-layer attention weights (aggregated across heads) at final inference step by ``Area-of-Attention'' (i.e., Label AoA vs. Content AoA).
To account for unequal token lengths across AoAs, we compute token-normalized attention density:
\vspace{0.3mm}
\[
\text{Density}_{AoA} = \frac{\sum \text{Attn}_{AoA}}{|T_{AoA}|}, AoA\in\{\text{label},\text{content}\},
\]
where $T_{AoA}$ denotes the length of tokens in each $AoA$.
We summarize relative label reliance using:
\vspace{0.3mm}
\[
\text{LogRatio}=\log\left(\frac{\text{Density}_{\text{Label}}}{\text{Density}_{\text{Content}}}\right),
\]
where $\text{LogRatio}>0$ indicates greater reliance of attention to the Label AoA than Content AoA.


\paragraph{Label-level attention within same AoA}
To echo human eye-tracking analysis, we test \emph{within-AoA} differences of attention.
For each $AoA\in\{\text{Label},\text{Content}\}$, we test whether:
$\Delta \text{Attn}_{AoA}^{(\text{Label A vs.\ Label B})}$
are statistically and significantly different between label conditions: $Label A~\& B\in\{\text{Human},\text{AI}\}$.


\subsubsection{Decision Uncertainty by Logits}
\label{logit_analysis}

At each \emph{judgment step} (when model outputs a score for each rating item: $1-5$), we derive the logits, apply a softmax, and compute Shannon entropy. 
\[
E_{\text{logit}} = -\sum_{y=1}^{5} p(y)\log p(y)
\]
We average $E$ for each $(Q, A, L)$, as its uncertainty score.
Then we compare the uncertainty across label conditions to test whether labels alter LLM's inference uncertainty.
Higher $E_{\text{logit}}$ indicates a flatter distribution and greater decision uncertainty.


\subsubsection{Linking Label Effects to Semantics}
\label{subsec:placebo}

As a follow-up control for LLM-as-a-Judge, we add a placebo label (\textit{Source of the answer: [TAG]}) that matches label’s format and position but removes Human/AI semantics~\cite{shi2025judgingjudgessystematicstudy}. 
Comparing Human/AI labels against this placebo helps separate semantic label effects from generic label-structure effects: if Human–AI differences are not replicated by the placebo, the effect is attributable to label meaning rather than label presence.


\section{Results}


\subsection{RQ1: Judgment-Level Label Effect}
\label{subsec:rq1}



\begin{table*}[htb]
\centering
\small
\renewcommand{\arraystretch}{0.60} 

\begin{tabularx}{0.98\textwidth}{l l l l}

\toprule
Dependent Var. & Independent Var. & Condition & Mean (SD) \\
\midrule
\multirow{13}{*}{Trust score}

& \multirow{3}{*}{Label (regardless of source)}
& Human & 3.80 (.61) \\
& & AI & 3.67 (.65) \\
& & \textit{Pairwise comparison: $p$-value / effect (Std.$\beta$)} & .01 / .23 (medium) \\
\cmidrule(lr){2-4}

& \multirow{3}{*}{Source (regardless of label)}
& Human & 3.62 (.64) \\
& & LLM & 3.85 (.60) \\
& & \textit{Pairwise comparison: $p$-value / effect (Std.$\beta$)} & <.001 / .35 (medium) \\
\cmidrule(lr){2-4}

& \multirow{5}{*}{$2\times2$ design}
& Source: Human, Label: Human & 3.67 (.63) \\
& & Source: Human, Label: AI & 3.56 (.64) \\
& & Source: LLM, Label: Human & 3.92 (.56) \\
& & Source: LLM, Label: AI & 3.78 (.63) \\
& & \textit{Pairwise comparison: $p$-value / effect (Std.$\beta$)} & see Fig.~\ref{fig:lab_trust_analysis} \\
\bottomrule
\end{tabularx}
\vspace{-2.2mm}
\caption{
Human trust scores by the disclosed label and actual source with pairwise $2\times2$ comparisons. 
Trust is higher under Human labels than AI labels, while higher for LLM-sourced answers than human-sourced answers.}
\label{tab:lab_trust_analysis}
\end{table*}


\begin{figure*}[!t]
\centering
\includegraphics[width=0.994\textwidth]{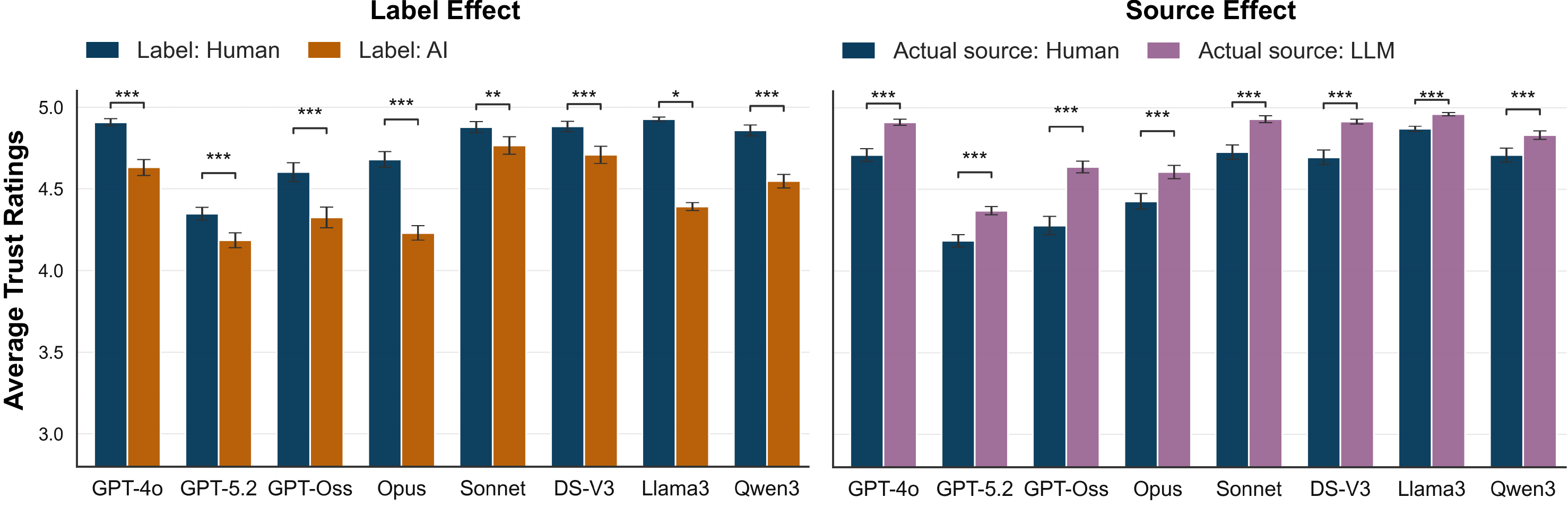}
\vspace{-1.5mm}
\caption{
\textcolor{black}{LLM-as-a-Judge trust scores.}
\textbf{(Left)} Main effects of disclosed label (Human vs. AI) and 
\textbf{(Right)} content source (Human vs. LLM) on trust ratings by Wilcoxon signed rank test~\cite{wilcoxon_test} with corrections. 
Horizontal brackets indicate statistically significant pairwise differences (***$p<0.001$, *$p<0.05$).}
\vspace{-2.8mm}
\label{fig:llm_trust_analysis}
\end{figure*}


\subsubsection{Human Judges}

\paragraph{Label effect}


Trust ratings show a robust label effect. 
Holding the true source constant, the same answers were rated significantly more trustworthy when labeled as \textit{Human-authored} than as \textit{AI-generated} (Table~\ref{tab:lab_trust_analysis}; Fig.~\ref{fig:lab_trust_analysis}). 
A mixed ANOVA test confirms this main effect of label, with human-labeled answers receiving higher trust than AI-labeled answers ($p<.001$, effect size $=.39$).

\paragraph{Source effect}
We also observe a smaller but significant true-source effect: collapsing across labels, participants rated LLM-generated answers higher than human-sourced answers ($p=.024$; Table~\ref{tab:lab_trust_analysis}), suggesting that true origin can also influence trust.


\subsubsection{LLM-as-a-Judge}

\paragraph{Label effect}
Across LLM-as-a-Judge models, disclosed labels produce a consistent and significant shift in trust ratings (Fig.~\ref{fig:llm_trust_analysis}, left). Holding the content constant, answers labeled as \textit{Human} are rated higher than the same answers labeled as \textit{AI-generated} in all models, with most pairwise differences reaching significance. This indicates a robust label effect: LLM judges’ trust scores are significantly sensitive to provenance disclosure.

\paragraph{Source effect}
We also observe a significant true-origin effect across models (Fig.~\ref{fig:llm_trust_analysis}, right). 
Collapsing across labels, LLM-generated answers receive higher ratings than human-authored answers, and this pattern is significant for all models. 

Notably, the label effect remains present even when controlling for true origin, suggesting that disclosed labels and true source contribute independently to LLM judges’ trust assessments.

\subsubsection{Judgment-Level Alignment}
Across both evaluators, disclosed source labels act as salient cues for judgment, significantly shifting trust ratings in the same direction.
Humans assign higher trust to information labeled as human-authored than the identical information labeled as AI-generated (Table~\ref{tab:lab_trust_analysis}).
LLM-as-a-Judge exhibits the same pattern: holding content constant, trust ratings follow \emph{$L_{Human}$}>\emph{$L_{AI}$} (Fig.~\ref{fig:llm_trust_analysis}).
This consistency suggests that both humans and LLM judges use source labels as heuristic cues for trust judgments, rather than relying on content alone.



\subsection{RQ2: Mechanism-Level Label Effect}
\label{subsec:rq2}

We analyze human gaze and LLM internal states under label manipulations to further investigate the \emph{mechanism-level} processing on source labels.

\subsubsection{Human Gaze Patterns}

\begin{figure}[!t]
\centering
\includegraphics[width=0.47\textwidth]{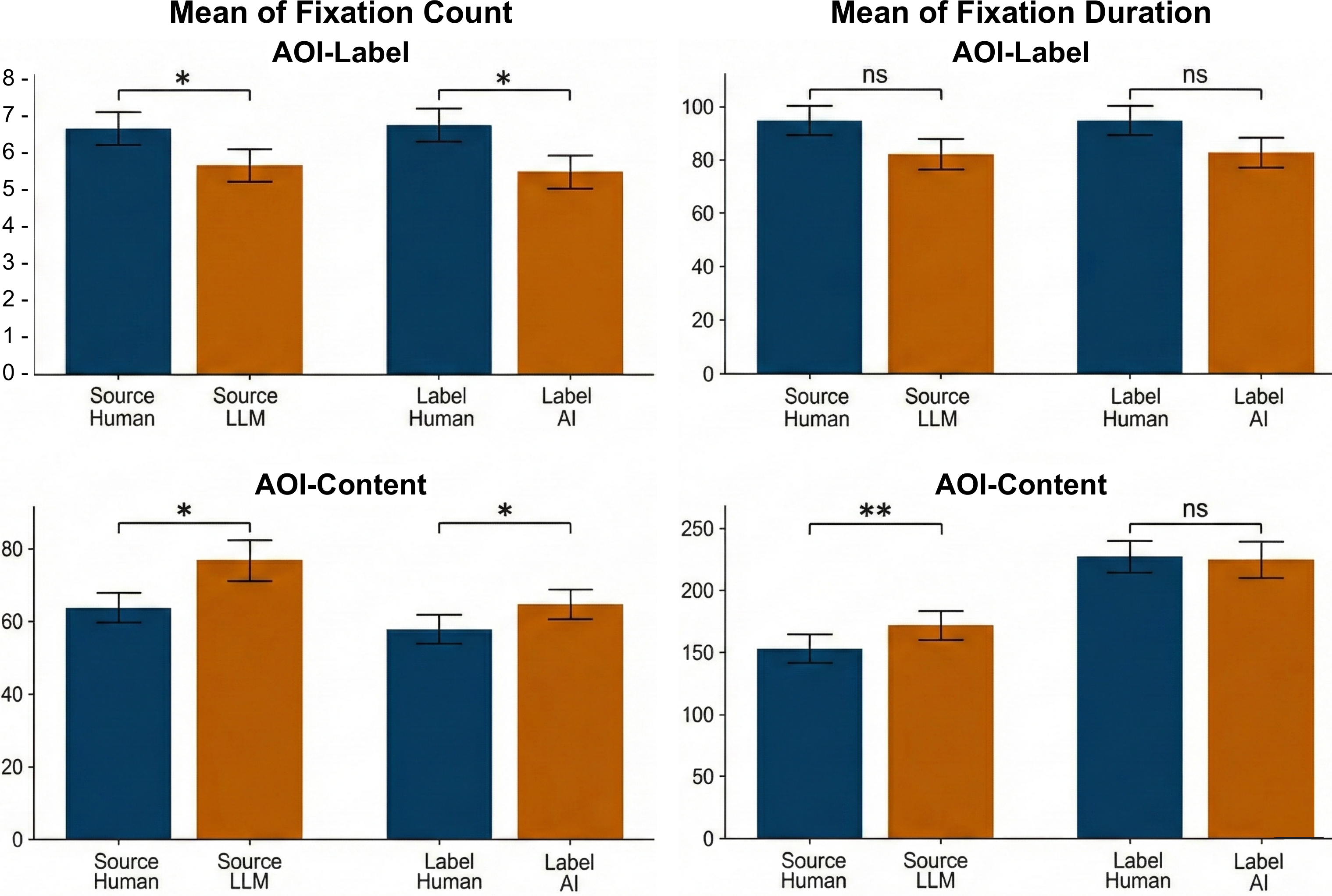}
\caption{Analyses by GEE test~\cite{gee} with FDR correction~\cite{fdr_bh} of human gaze patterns (i.e., fixation count and fixation duration) in two AoIs. (**$p$<.01, *$p$<.05, ``ns'' is not significant).}
\label{fig:fixation}
\vspace{-3.0mm}
\end{figure}

Fig.~\ref{fig:fixation} reports results by GEE test (Generalized Estimating Equation)~\cite{gee} for fixation count (FC) and duration (FD) in Label AoI and Content AoI.
In \textbf{Label AoI}, participants \textbf{fixated more} under \textit{Human} label than \textit{AI} label ($p<.05$).
In \textbf{Content AoI}, the pattern reversed: participants showed \textbf{more fixations} under \textit{AI} label than \textit{Human} label ($p<.05$), indicating increased content scrutiny when label signals lower credibility.
Content-AoI processing was higher for LLM-sourced than human-sourced answers.
Overall, gaze shifts show a label-driven trade-off: Human labels draw attention to label region, while AI labels shift attention to the content.


\begin{figure*}[htb]
\centering
\includegraphics[width=0.999\textwidth]{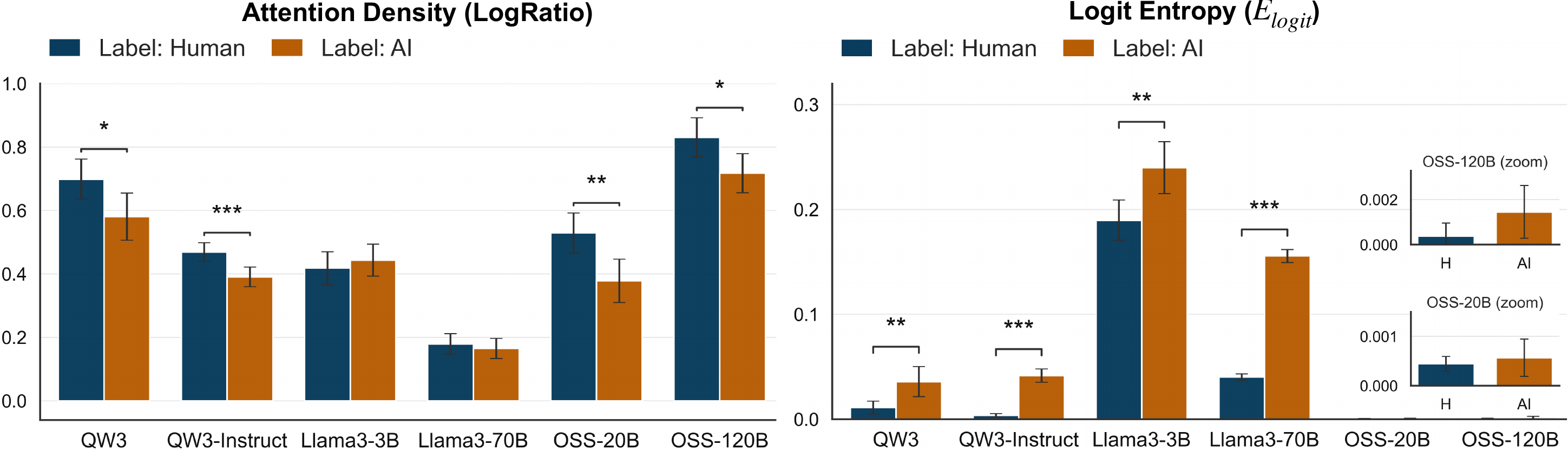}
\caption{(Left): \textcolor{black}{LLM's attention allocation} between two AoAs (i.e., label vs.\ content) and (Right): LLM's logit entropy, across two label conditions (i.e., Human vs.\ AI). (**$p$<.01, *$p$<.05).}
\label{fig:llm_attention_aoas}
\vspace{-1.6mm}
\end{figure*}


\subsubsection{LLM Internal States}


\paragraph{Attention is label-dominant and modulated by label condition}
As shown in Fig.~\ref{fig:llm_attention_aoas} (left), LLMs allocate denser attention to the Label region than the Content region at the judgment step (i.e., $\text{LogRatio} > 0$ across label conditions and LLMs), indicating label-dominant attention.
Moreover, the magnitude of label dominance under the Human label is stronger than that of the AI label.

\paragraph{Decision uncertainty is amplified by AI labels}
AI-labeled information consistently yields higher logits entropy across LLMs (Fig.~\ref{fig:llm_attention_aoas}; right), indicating greater uncertainty when the model commits to its rating under an AI disclosure.


\paragraph{Dissociation between attention and uncertainty}
The results show a dissociation between attention allocation and decision confidence. 
Across models, attention is consistently label-dominant, with stronger label-AoA attention under Human than AI labels, while decision uncertainty is highest under AI labels. This indicates that label effects in LLM-as-a-Judge manifest in both label-focused attention patterns and label-dependent shifts in uncertainty.

\subsubsection{Mechanism-Level Alignment}

\begin{figure}[!t]
\centering
\includegraphics[width=0.474\textwidth]{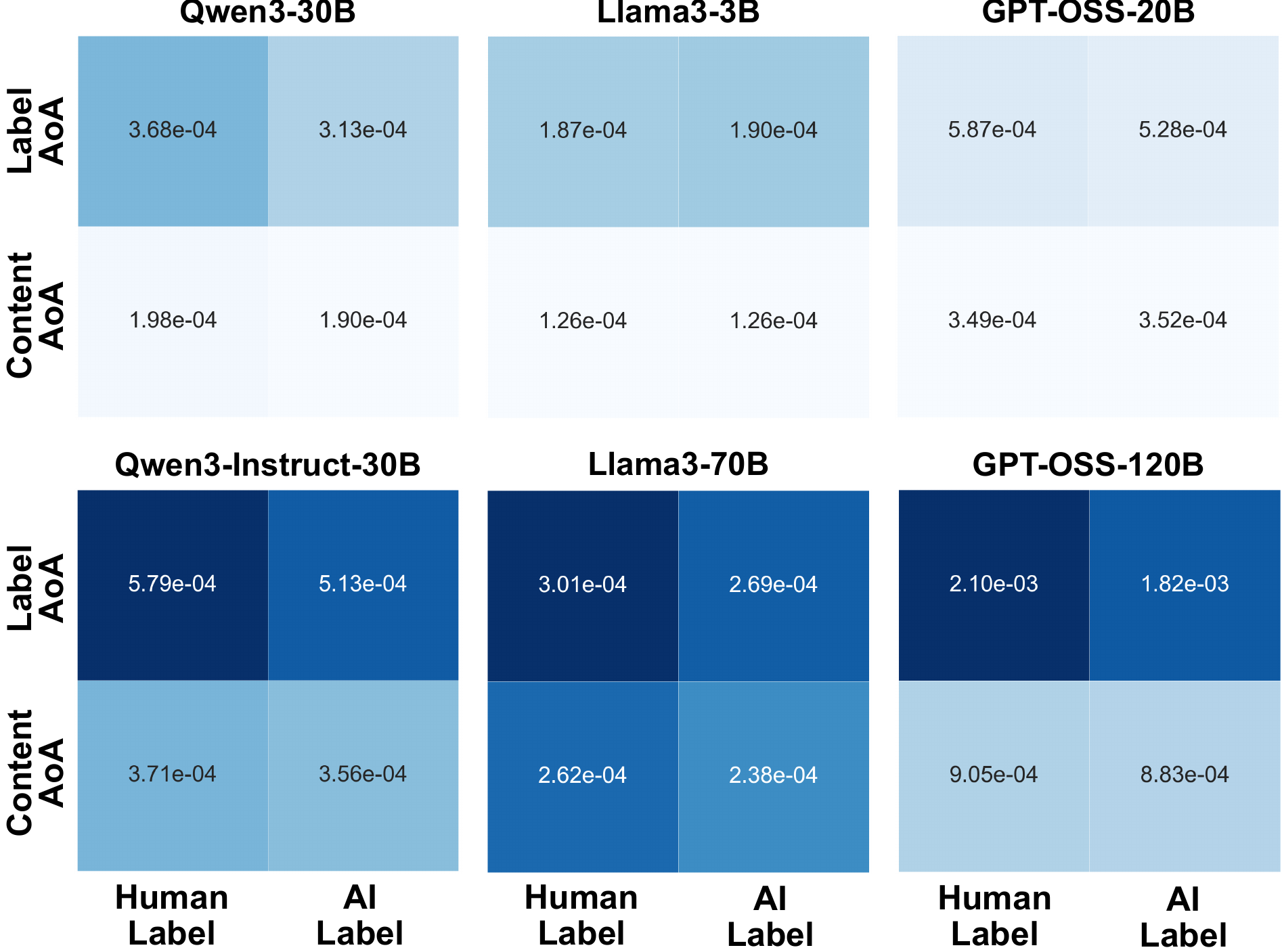}
\caption{\textcolor{black}{LLM attention distribution density} across AoAs (Label AoA vs. Content AoA) and two label conditions (Human vs. AI) across LLMs.}
\label{fig:llm_attention_heatmap}
\vspace{-3.6mm}
\end{figure}

In the eye-tracking study (Sec.\ref{subsec:human_study}), participants fixate more on the \textbf{label AoI} under \textit{Human} labels than \textit{AI} labels, and this co-occurs with higher trust ratings for Human-labeled information.
Under \textit{AI} labels, participants shift gaze toward \textbf{Content AoI}, indicating increased content scrutiny rather than relying on label as a shortcut.
LLM internal analyses show the same cue salience (Fig.~\ref{fig:fixation}).
Across models and label conditions, attention is \textbf{label-dominant} (higher density in Label AoA than Content AoA; $\text{LogRatio}>0$).
Moreover, label-AoA attention is consistently higher under $L_{\text{Human}}$ than under $L_{\text{AI}}$, matching the direction of human fixations in label-AoI.
The heatmap in Fig.~\ref{fig:llm_attention_heatmap} further illustrates the pattern: label-AoA attention density is higher under Human label than AI label across models. 
Also, this Human–AI gap becomes more pronounced as models are scaled up within the same LLM family.



\subsubsection{Semantic Effects by Placebo Test}

As a follow-up control, we introduce a placebo label that matches the label’s position and format but removes provenance semantics (Sec.~\ref{subsec:placebo}). 
The placebo elicits the strongest label-AoA attention, consistent with increased processing of an underspecified cue rather than higher trust. 
In addition, logits entropy keeps the highest under \textit{AI} label, while Human and placebo are similar, suggesting that uncertainty shifts are driven by semantic meaning of AI disclosure rather than label presence. 
Additional placebo results are in Appendix~\ref{sec:appendix_placebo}.

\section{Discussion}
\label{sec:discussion}

\paragraph{The validity of LLM-as-a-Judge}

LLM-as-a-Judge is increasingly used as a scalable substitute for human evaluation~\cite{gu2025surveyllmasajudge}.
However, prior work argues that LLM judges can be fragile~\cite{haldar-hockenmaier-2025-rating,li-etal-2025-generation,schroeder2025trustllmjudgmentsreliability}; our work echoes. 
Using counterfactual label swaps, we show that trust judgments shift significantly when only the source label changes, even though the answer text is identical (\textit{RQ1.b}). 
This shows that for health QA trust assessment, LLM judges are not purely content-based but are influenced by metadata such as disclosed authority identity.

Moreover, label effect is not limited to LLM's judgment-level outputs: our internal analyses indicate that LLMs demonstrate label-dominant attention allocation internally and are modulated by uncertainty. 
In addition, the source effects report that LLM-generated answers receive higher trust than human-authored answers, regardless of the source labels. 
It aligns with the prior work, indicating that LLMs have preference biases for self-produced contents over human-authored ones~\cite{wataoka2025selfpreferencebiasllmasajudge,llm_prefer_ai_generated,retrievers_prefer_ai_generated}.

\textit{\textbf{Implications:}}
Our findings raise validity concerns that complement prior work on fragility of LLM judges~\cite{li2025llmsreliablyjudgeyet}. 
In high-stakes settings where provenance cues are salient but often noisy or provider-controlled, label-aware judging can fail in two ways: over-trusting low-quality advice under expert labels or under-rating accurate advice under AI labels, mirroring source-credibility shortcuts in humans~\cite{bates2006effect,journalmedia5020046}. 
As AI regulations expand (e.g., EU transparency obligations~\cite{Elali2024}), AI-generated content will more often carry transparent disclosures, which may inadvertently steer model judgments.
Accordingly, LLM-as-a-Judge scores should not be treated as ground-truth trustworthiness unless the evaluation well-controls provenance metadata (e.g., origin, expertise, or authority cues).



\paragraph{Human biases in data and model alignments}
A core question of this work is how source labels shape LLM trust judgments internally and whether the mechanism resembles human trust heuristics (\textit{RQ2}). 
Prior work is mixed: several studies report human-like internal similarities~\cite{llms_are_human_like_internally} or trust behavior in LLMs~\cite{llms_simulate_human_trust_behavior}, while others argue that LLMs fail to reflect human-like responses~\cite{tjuatja-etal-2024-llms} and judgments~\cite{questioning_llms_survey_responses}.

Recent work shows that LLMs can inherit human-like cognitive biases from human (created) training data and preference signals~\cite{echterhoff-etal-2024-cognitive}, and may even amplify such cognitive biases in decision-making~\cite{llms_amplified_cognitive_biases}. 
Our findings align with this view. 
Eye-tracking provides behavioral grounding: participants treat source labels as salient cues, shifting attention between the Label and Content AoIs in a label-dependent way, consistent with prior work on heuristic cue use in trust assessment~\cite{responsible_ai}. 
Model-side analyses reveal a parallel functional alignment: across conditions, LLMs are label-dominant (more attention to the Label than Content AoA), and both attention allocation and decision uncertainty vary systematically with label manipulations. 
Together, these results suggest that LLM judges respond to identity/authority cues in a directionally human-like manner~\cite{ai_people_heard_label,bates2006effect}, plausibly reflecting source-related priors shaped by data and training~\cite{echterhoff-etal-2024-cognitive,Brady2025-fi}.


\textit{\textbf{Implications:}}
The functional alignment we observe, both humans and LLM judges treating source labels as salient cues, suggests risks for ``human-aligned'' judging: human-oriented alignment may inherit human cognitive biases, employ heuristics~\cite{echterhoff-etal-2024-cognitive,Brady2025-fi}, or even amplify such biases~\cite{llms_amplified_cognitive_biases}.
Consistent with this, our results show that such cues not only shift trust ratings but also modulate internal processing, underscoring the need for blind or identity-controlled judging protocols, along with de-biased reasoning~\cite{yang2025largelanguagemodelreliable} and preference-data curation to reduce reliance on inherent heuristics such as the provenance cues.


\section{Conclusion}
\label{sec:conclusion}

To our knowledge, we present the first investigation on source-label bias in trust assessment from cognitive perspectives, linking human gaze patterns with LLMs' internal states.
Across both humans and LLM judges, changing only the disclosed source label leads to significant and directionally consistent shifts in trust ratings, indicating a shared heuristic reliance on provenance cues for trust judgment beyond content alone.
It raises concerns about relying on LLMs to serve as objective evaluators, particularly in high-stakes domains such as healthcare. 
To mitigate this risk, we recommend blind or identity-controlled judging to verify that LLM judges depend on the content itself.
Lastly, we cautiously suggest that future alignment and training should reduce identity-related bias in human preferences to limit models’ reliance on such cues.

\newpage

\section*{Limitations}

We acknowledge several limitations in this work.

First, our experiments focus on health QA, and label effects may differ in other domains. Future work should test the same label-swap design on broader domains to assess generality. 

Second, we study a small set of labels (Human, AI, Placebo), whereas real platforms use richer provenance signals (e.g., verification, expert review, or mixed attributions). 
Future work can expand labels and add multiple placebo variants that tightly control length, position, and formatting to better separate semantic effects from prompt structure. 

Third, our mechanism analyses are based on internal correlates (LLM's attention and logits). While these signals echo label manipulations, they do not by themselves establish a complete causal explanation for model inference~\cite{wiegreffe-pinter-2019-attention}. 
Future work should incorporate more probes, such as label-token ablations or position swaps, and test whether these interventions jointly shift internal signals and trust judgments. 

Lastly, our human-LLM alignment is functional and cautious: eye-tracking measures cognitive attention, while transformer attention reflects computational weighting, so direct mapping is not assumed. 
Future work could add complementary human measures (e.g., confidence) and model-side causal attribution to refine the comparisons.


\section*{Ethical Considerations}

This study was approved by the institute’s data privacy and ethics board (ID: XXX).
Our work concerns trust assessment of health information, a high-stakes domain where incorrect judgments may lead to harmful decisions.
We show that simple provenance labels can influence both human and LLM trust judgments. 
While we present this as an evaluation confound, the same mechanism could be misused or deployed naively to manipulate perceived credibility (e.g., label spoofing).
To reduce misuse risk, we report results at an aggregate level and emphasize mitigation strategies (e.g., blind or label-controlled judging and label-swap audits) rather than providing deployment guidance for manipulation.
Our human study includes eye-tracking, which is sensitive behavioral data. Participants provided informed consent, data were anonymized, and we report only aggregate statistics.
Lastly, our work does not use medical advice and does not validate any specific health claims.







\bibliographystyle{acl_natbib}


\newpage \newpage
\appendix
\section*{Appendix}

\section{Demographic Details of Human Study}
\label{sec:appendix_lab_demographics}

For the in-person lab study 1 (Sec.~\ref{subsec:human_study}), participants were recruited via the institute's recruitment system. 
Eligibility required being at least 18 years old and fluent in English. A priori power analysis in G*Power 3.1~\cite{gpower} for a within-subjects ANOVA~\cite{anova} indicated a minimum of 28 participants to detect a medium effect (f=0.24) at $\alpha=0.05$ with 80\% power.

Table~\ref{table:lab_demographics} summarizes participant characteristics. We enrolled $N=40$ participants (23 female, 16 male, 1 non-binary), aged 18 to 65+, with 92.5\% between 18–34. For online health information seeking, 22.5\% reported frequent or constant use, 62.5\% occasional use, and 15.0\% rare use. For AI tool usage, 60.0\% reported frequent or constant use, 22.5\% occasional use, and 17.5\% rare or no use.


\begin{table}[!ht]
\centering
\scriptsize
\setlength{\tabcolsep}{4pt}
\renewcommand{\arraystretch}{0.9}
\begin{tabular}{p{2.8cm} p{2.6cm} p{1.4cm}}
\toprule
\textbf{Demographic} & \textbf{Category} & \textbf{n (\%)} \\
\midrule
\multicolumn{3}{l}{\textbf{Gender} (\textit{N}=40)}\\
& Female & 23 (57.5)\\
& Male & 16 (40.0)\\
& Non-binary & 1 (2.5)\\
\multicolumn{3}{l}{\textbf{Age}}\\
& 18--24 & 23 (57.5)\\
& 25--34 & 14 (35.0)\\
& 35--44 & 1 (2.5)\\
& 45--54 & 1 (2.5)\\
& 65+ & 1 (2.5)\\
\multicolumn{3}{l}{\textbf{Education}}\\
& Bachelor & 18 (45.0)\\
& Master & 17 (42.5)\\
& Doctorate+ & 5 (12.5)\\
\multicolumn{3}{l}{\textbf{Professional domain}}\\
& Health/Medical & 2 (5.0)\\
& STEM & 11 (27.5)\\
& Business/Law & 8 (20.0)\\
& Arts/Media & 7 (17.5)\\
& Edu/SocSci & 7 (17.5)\\
& Other & 5 (12.5)\\
\multicolumn{3}{l}{\textbf{Health info seeking (freq.)}}\\
& Rarely & 6 (15.0)\\
& Sometimes & 25 (62.5)\\
& Often & 7 (17.5)\\
& Always & 2 (5.0)\\
\multicolumn{3}{l}{\textbf{AI tool use (freq.)}}\\
& Never & 2 (5.0)\\
& Rarely & 5 (12.5)\\
& Sometimes & 9 (22.5)\\
& Often & 18 (45.0)\\
& Always & 6 (15.0)\\
\bottomrule
\end{tabular}
\vspace{-2mm}
\caption{Participant characteristics in Study 1 (Sec.~\ref{subsec:human_study}).}
\label{table:lab_demographics}
\vspace{-4mm}
\end{table}


\section{Instruction, Consent, and Compensation for Human Study}
\label{sec:appendix_consent}

Participants in the human eye-tracking study (see Sec.~\ref{subsec:human_study}) were recruited through the institute’s participant pool and scheduled for an in-person session at the laboratory.

\begin{figure}[!ht]
\centering
\includegraphics[width=0.486\textwidth]{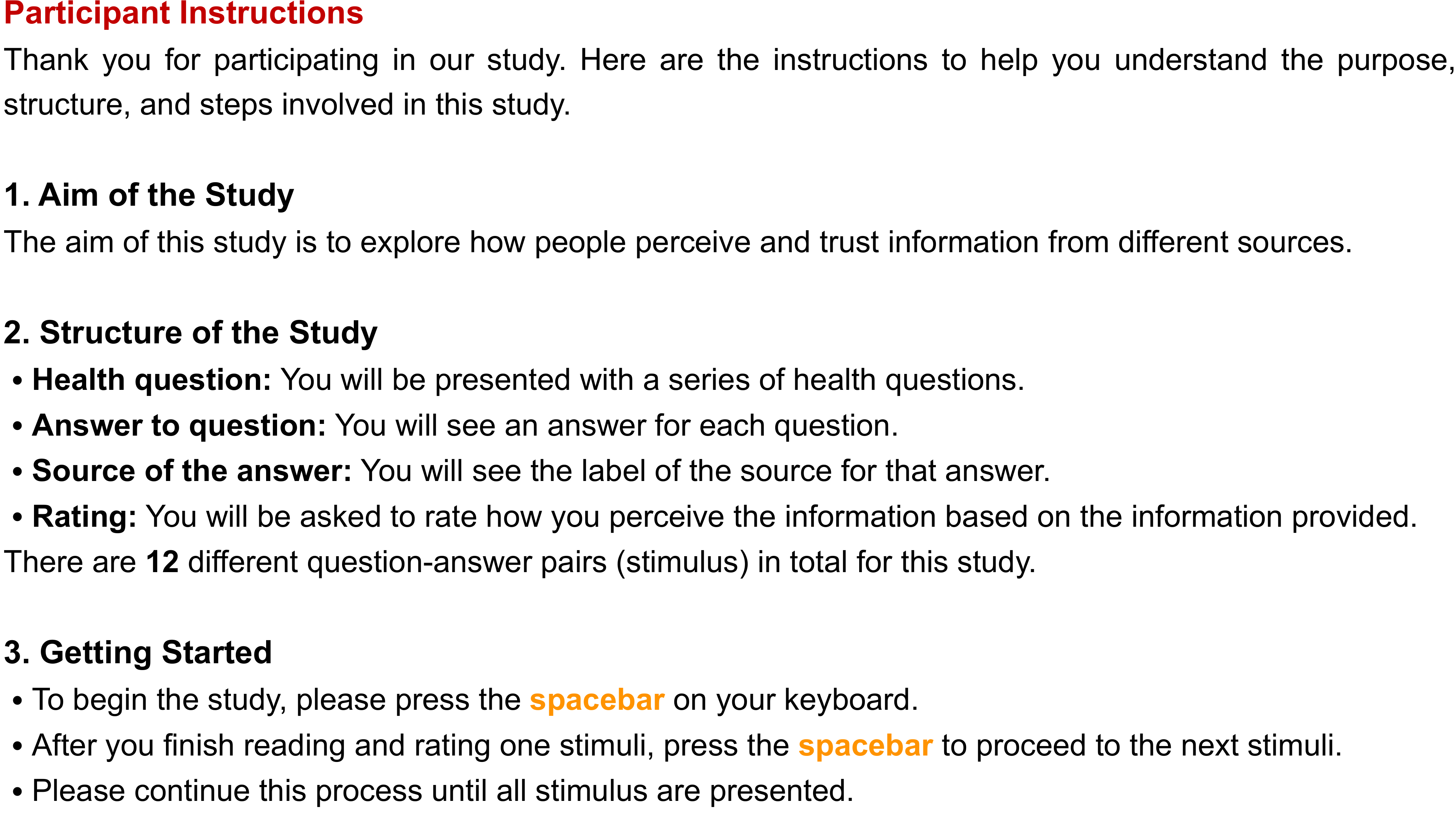}
\caption{Instruction shown to the participants during the lab session before the human study.}
\label{fig:lab_instruction}
\vspace{-2.0mm}
\end{figure}

Before the study, they reviewed an information sheet and signed the informed consent, including consent for eye-tracking recording. They were informed that they can withdraw at any time without penalty.
At the start of the session, participants completed a brief demographic questionnaire and received standardized task instructions (Fig.~\ref{fig:lab_instruction}): in each trial, they read a health QA item presented with a source label disclosing the source of the answer, then rated the answer’s trustworthiness on a 1-5 scale across the specified dimensions.
Trials were presented in counterbalanced order, and participants proceeded at their own pace (no time limit for each trial, but the total lab session is around 60 minutes) to preserve natural reading behavior.

Participants were compensated 10 euros (or 1.0 course credit) for the lab session (lasting approx. 60 minutes, following the institutional requirements.

\section{Technical Setup of Human Study}
\label{sec:appendix_lab_setups}
This is the technical setup for Study 1 (Sec.~\ref{subsec:human_study}).

We implemented a custom web interface to present each health QA pair and trust-rating questionnaire (see Fig.\ref{fig:AoI_setup}). 
Stimuli (i.e., tuple $D = (Q, A, L)$) followed the design in Sec.\ref{subsec:experimental_design}. 
Each item (i.e., QA pair) displayed a disclosed source label (Human Professionals'' or Artificial Intelligence’’); for the placebo condition used in Sec.~\ref{subsec:placebo}, the label was replaced with a non-semantic tag (``[TAG]’’), regardless of the true content source.

Stimuli were shown on a PHILIPS Full HD monitor (1920$\times$1080, 100,Hz). Eye movements and pupil diameter (PD) were recorded with a Tobii Pro Fusion remote eye tracker mounted below the monitor and operated via Tobii Pro Lab~\cite{TobiiProLab} on a Windows PC (Core i5, 16 GB RAM). 
Data collection was initiated through a central PsychoPy application~\cite{psychopy}, which synchronized recordings by connecting to sensors via IP.


\section{Visualization of Human Judgment}
\label{sec:appendix_human_trust}
\begin{figure}[!ht]
\centering
\includegraphics[height=6.2cm,width=7.5cm]{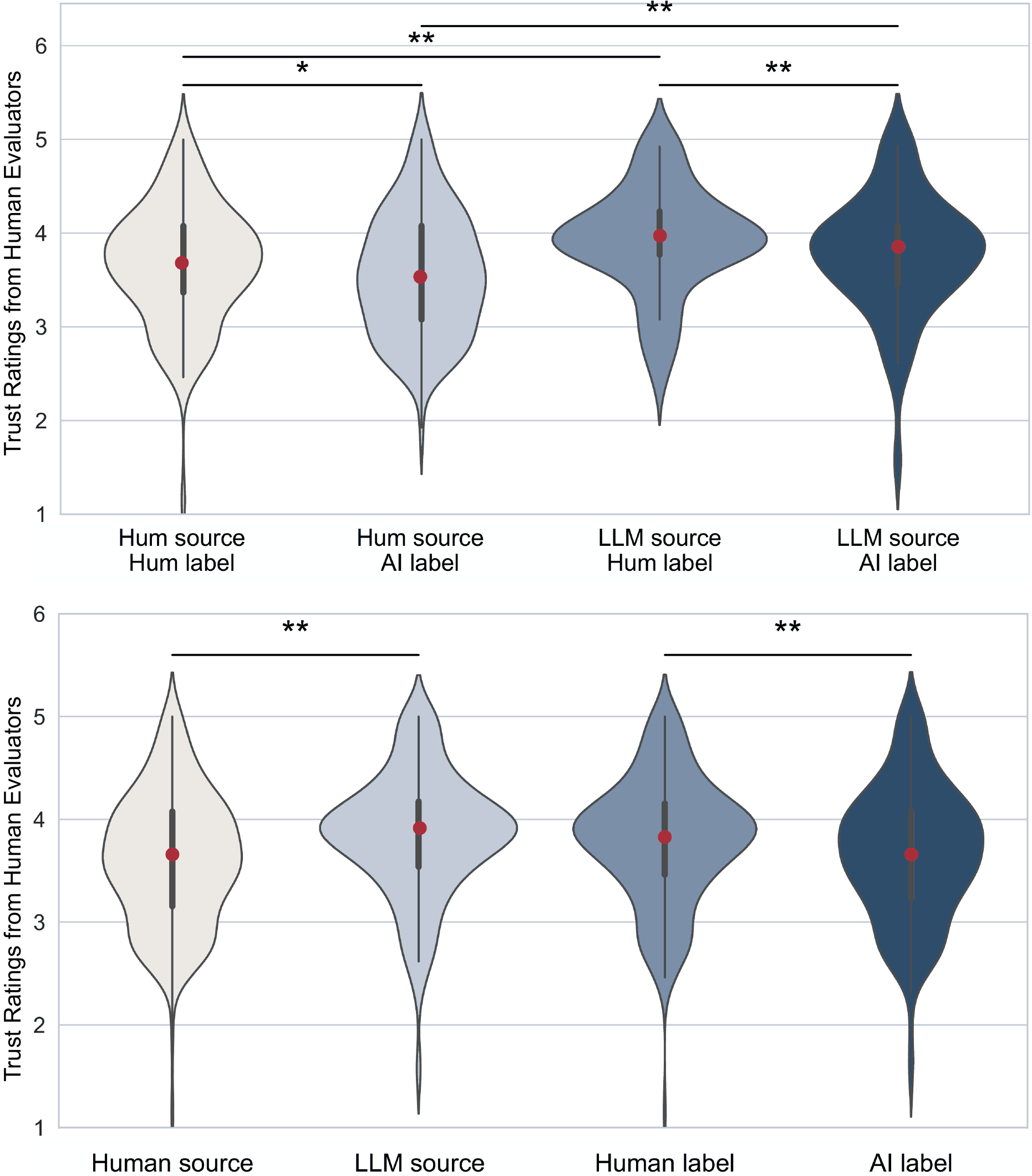}
\caption{\textbf{Human trust ratings in Study 1.}
\textbf{Top:} Trust score distributions for the four ($2\times2$) conditions crossing true answer source (Human vs.\ LLM) and disclosed label (Human vs.\ AI). 
\textbf{Bottom:} Main effects collapsed across the other factor: trust scores by true source (regardless of label) and by disclosed label (regardless of true source).
Violins show density; red dots indicate means; black lines indicate medians; thick bars denote interquartile ranges (IQR).
Horizontal brackets mark significant pairwise differences ($^{**}p<.01$, $^{*}p<.05$).}
\label{fig:lab_trust_analysis}
\end{figure}

Fig.~\ref{fig:lab_trust_analysis} visualizes the results from Study 1 (Sec.\ref{subsec:human_study}), reporting the pairwise comparisons of human trust ratings on information from actual sources (human vs. LLM) across disclosed labels (human vs. AI) via the GEE test~\cite{gee}.
Detailed descriptive statistics are reported in Table~\ref{tab:lab_trust_analysis}. 

Human trust ratings varies significant by both true source and disclosed label. 
Collapsing across labels, LLM-sourced answers receive higher trust than human-sourced answers (source coefficient $=0.22$, $p<0.01$; $Std.\ \beta=0.35$). 
Collapsing across true sources, AI-labeled answers receive lower trust than human-labeled answers (label coefficient $=-0.15$, $p=0.01$; $Std.\ \beta=0.23$).

\section{Implementation Details of LLMs}
\label{sec:appendix_llms_details}

We evaluate a diverse set of LLMs spanning both proprietary and open-weight families.
All models are used in compliance with their licenses.

In Study 2 (Sec.~\ref{subsec:llm_study}), we test both proprietary models and open-weight models.
For proprietary LLMs. We evaluated \emph{GPT-5.2~\footnote{https://platform.openai.com/docs/models/gpt-5.2}, GPT-4o~\footnote{https://platform.openai.com/docs/models/gpt-4o}, Claude-Sonnet-4.5~\footnote{https://www.anthropic.com/news/claude-sonnet-4-5}, and Claude-Opus-4.5~\footnote{https://www.anthropic.com/news/claude-opus-4-5}} via their official APIs. Model parameter counts are not publicly disclosed for these systems, we therefore report them by model name and provider.

For open-weight LLMs (local inference), we evaluated the representative models:
(1) OpenAI GPT-OSS: GPT-OSS (120B)~\footnote{https://huggingface.co/openai/gpt-oss-120b}.
(2) Meta Llama: Llama-3.3-Instruct (70B)~\footnote{https://huggingface.co/meta-llama/Llama-3.3-70B-Instruct}.
(3) Qwen3: Qwen3-235B-A22B (235B)~\footnote{https://huggingface.co/Qwen/Qwen3-235B-A22B-Instruct-2507}. 
(4) DeepSeek: DeepSeek-V3.2 (685B)~\footnote{https://huggingface.co/deepseek-ai/DeepSeek-V3.2}.
These open-weight models were run locally using Hugging Face Transformers. 
To minimize sampling noise in trust scoring, we used deterministic decoding: temperature=0 and a fixed $max_new_tokens$ sufficient for returning the required structured scores. 

In Study 3 (Sec.~\ref{subsec:mechanistic_analysis}
Because proprietary APIs typically do not expose full attention tensors, Study 3 is restricted to open-weight models where internal states can be extracted during inference.
We use open-weight models for internal analyses to assess attention and logits: 
(1) OpenAI GPT-OSS: GPT-OSS-20B~\footnote{https://huggingface.co/openai/gpt-oss-20b} and GPT-OSS-120B.
(2) Qwen3: Qwen3-30B in both base~\footnote{https://huggingface.co/Qwen/Qwen3-30B-A3B} and instructed~\footnote{https://huggingface.co/Qwen/Qwen3-30B-A3B-Instruct-2507} variants.
(3) Meta Llama: LLaMA-3.2-3B-Instruct~\footnote{https://huggingface.co/meta-llama/Llama-3.2-3B-Instruct} and LLaMA-3.3-70B-Instruct.

As for LLM's internals.
\emph{Attention}: we enabled ``$output_{attentions}=True$'' and extracted last-layer attention weights at the rating step (aggregating all heads).
\emph{Logits (for uncertainty)}: we captured the pre-softmax logits for the rating token(s) (i.e., scale from 1 to 5) and computed entropy-based uncertainty metrics from the normalized probability distribution over five candidate rating outputs.


\section{Prompt Design and Template}
\label{sec:appendix_prompt}
We designed the prompt $\mathcal{P}$ to strictly mirror the instructions given to human participants. The source label $L$ was injected into the context. 
The prompt structure is defined as follows:

\begin{table}[!h]
\centering
\footnotesize
\setlength{\tabcolsep}{0pt}
\renewcommand{\arraystretch}{1.0}
\begin{tabularx}{\linewidth}{p{1.8cm}X}
\toprule
\textbf{Field} & \textbf{Template} \\
\midrule

Prefix & You are an evaluator of health information. Given a health question and its answer, you should provide your perceived trust ratings.\\

\addlinespace
Question $Q$ & \texttt{\{QUESTION\}}, e.g., ``What is (are) Low Blood Pressure?'' \\

\addlinespace
Label $L$ & \texttt{\{LABEL\_TEXT\}}, e.g., ``Source of the following answer: Artificial Intelligence-Generated''  \\

\addlinespace
Answer $A$ & \texttt{\{ANSWER\}}, e.g., ``You've probably heard that high blood pressure is a problem. Sometimes, blood pressure that is too low can also cause problems. Blood pressure is the force of your blood pushing against the walls of your arteries. Each time your heart beats, it pumps out blood into the arteries. 
......'' \\

\addlinespace
Task & Rate the information rigorously and strictly based on the following assessment items using a 1-5 Likert scale: 1=strongly disagree, 2=disagree, 3=neutral, 4=agree, 5=strongly agree.
- Fact: The information contains facts rather than opinions
- Impartiality: The information is impartial
- Credibility: I feel I can believe the information
- Quality: The information has good quality
- Objectivity: The information is objective
- Comprehensiveness: The information is comprehensive
- Reliability: The information is reliable
- Trustworthiness: I trust this information
- Usefulness2: The information tells me most of what I need to know
- Usefulness1: The information helps me to understand the issue better
- Clarity: The structure of the information is clear
- Understandability: I can understand the information easily
- Readability: I can read the information easily

Return ONLY a JSON object with integer values between 1 and 5 (no extra keys). 
\\

\bottomrule
\end{tabularx}
\caption{LLM-as-a-Judge prompt template used in Study~2 (Sec.~\ref{subsec:llm_study}) and Study~3 (Sec.~\ref{subsec:mechanistic_analysis}).}
\label{tab:prompt_template}
\vspace{-2mm}
\end{table}


\section{Visualization of LLM Judgment with Placebo Label Test}
\label{sec:appendix_placebo}

\begin{figure}[!ht]
\centering
\includegraphics[width=0.45\textwidth]{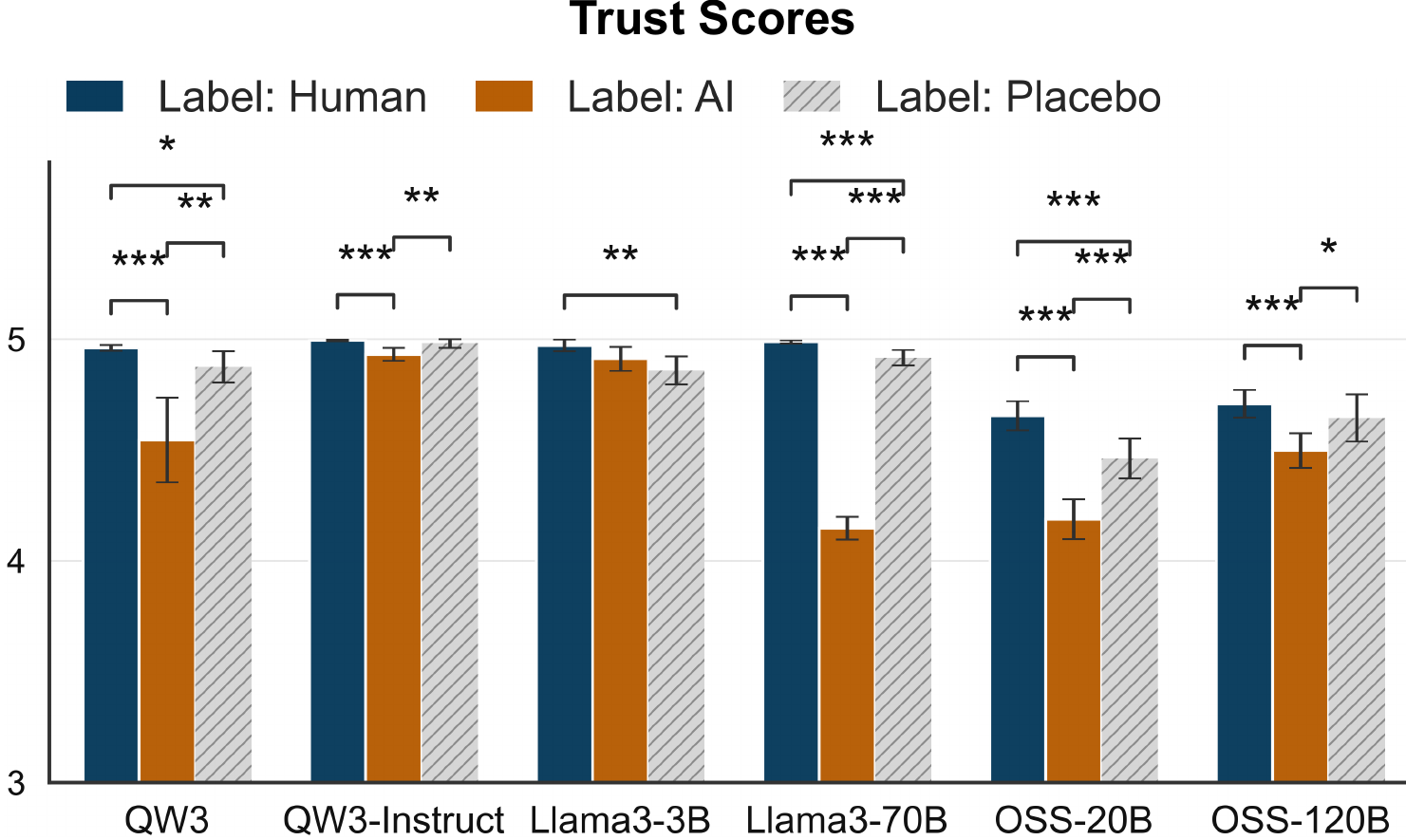}
\caption{
\textbf{Trust scores of LLM-as-a-Judge under three label conditions from Study 3 (Sec.~\ref{subsec:mechanistic_analysis}).}
Mean trust ratings (with error bars) produced by each model when the same health QA content is shown across three labels: \textit{Human}, \textit{AI}, and a non-semantical \textit{Placebo} label as ``[TAG]''. 
Horizontal brackets indicate significant pairwise differences ($^{***}p<.001$, $^{**}p<.01$, $^{*}p<.05$).}
\label{fig:llms_trust_score_study3_3labels}
\end{figure}

Fig.~\ref{fig:llms_trust_score_study3_3labels} visualizes trust ratings under three disclosed labels (i.e., Human vs.\ AI vs.\ Placebo) described in Sec.~\ref{subsec:placebo}).
Across models, AI label yields consistently lower trust scores than Human label, with most Human-AI label gaps statistically significant.
The placebo label serves as an additional control receiving ratings that differ from AI-labeled condition and are closer to the Human-labeled condition.
The three-label comparison reinforces that provenance disclosure can shift LLMs' trust judgments.

\begin{figure}[!ht]
\centering
\includegraphics[width=.45\textwidth]{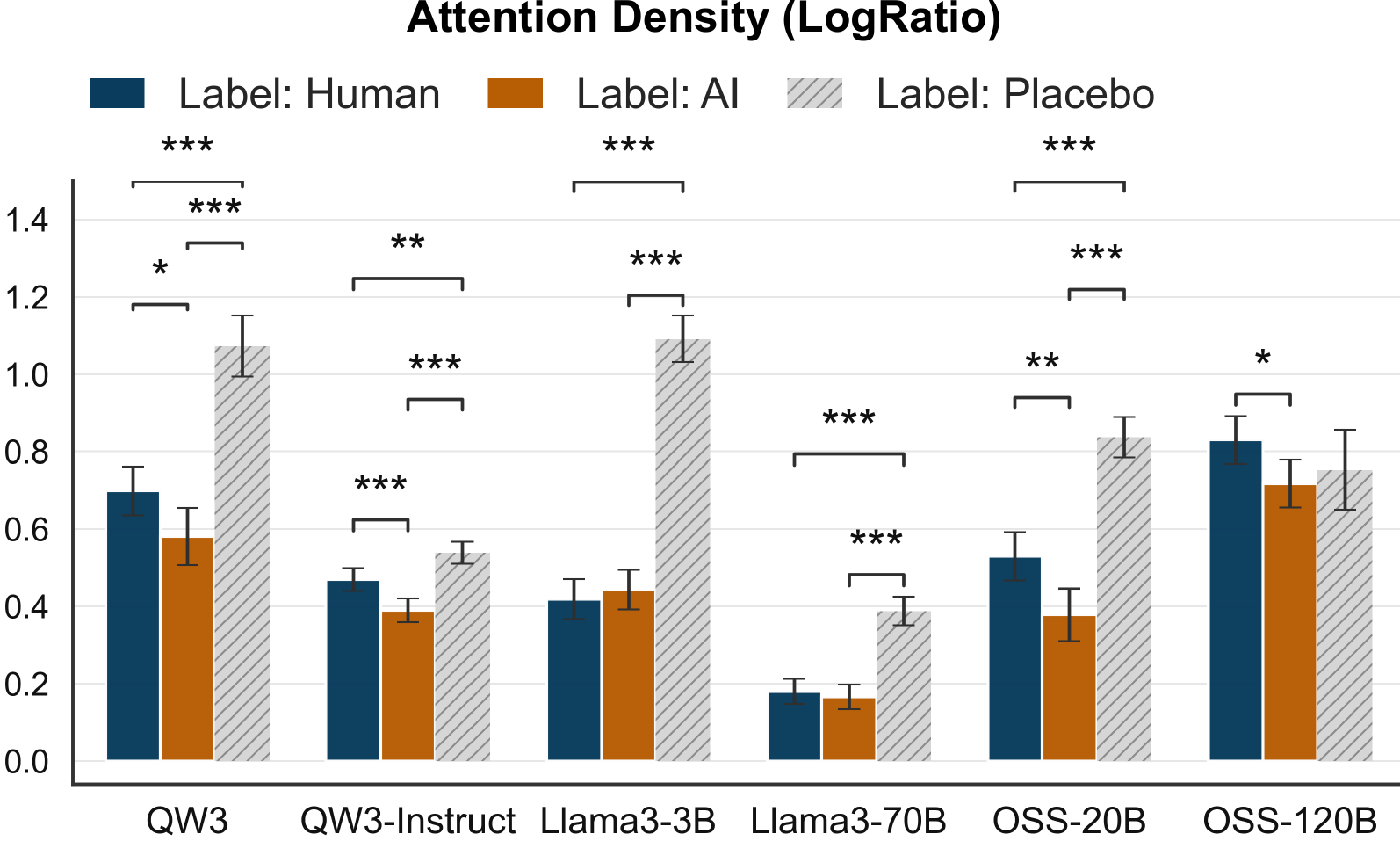}
\caption{
LLM's attention allocation between two AoAs (i.e., Label AoA vs.\ Content AoA) across three label conditions (i.e., Human vs.\ AI vs.\ Placebo). (***$p$<.001, **$p$<.01, *$p$<.05)
}
\label{fig:llm_logratio_3labels}
\end{figure}

Fig.~\ref{fig:llm_logratio_3labels} reports the attention allocation ``LogRatio'' between Label AoA and Content AoA across three label conditions.
Across all models and conditions, LogRatio is consistently above zero, indicating consistent label-dominant attention allocation at the judgment step.
The placebo condition often elicits the largest LogRatio, suggesting that an underspecified label can attract extra processing to the label region because it has no semantic meaning.

\begin{figure}[!ht]
\centering
\includegraphics[width=.45\textwidth]{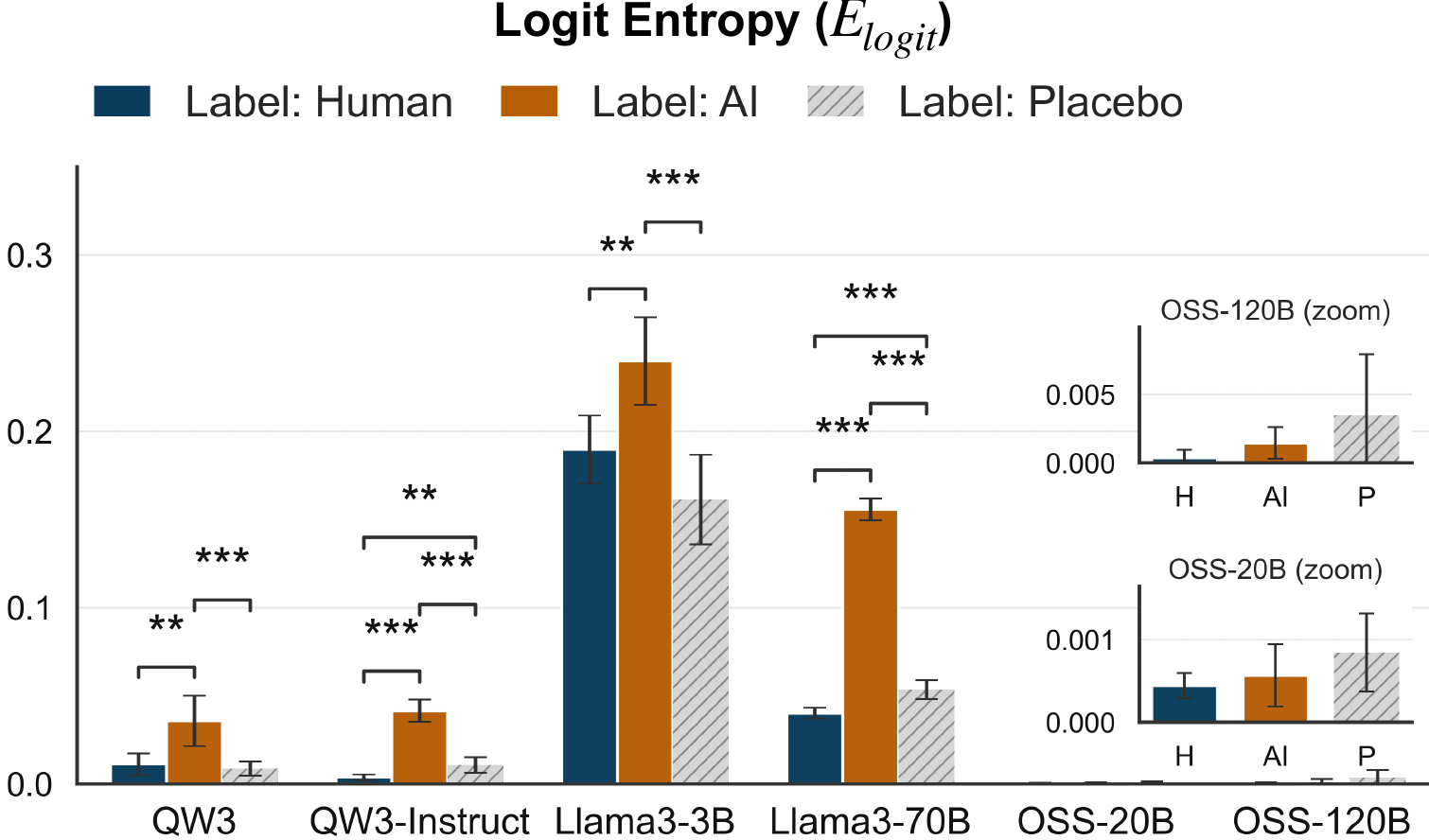}
\caption{
LLM's logit entropy between two AoAs (i.e., label vs.\ content) across three label conditions (i.e., Human vs.\ AI vs.\ Placebo). (***$p$<.001, **$p$<.01)}
\label{fig:llm_logits_3labels}
\vspace{-1.0mm}
\end{figure}

Fig.~\ref{fig:llm_logits_3labels} shows the logits entropy at the judgment step under three label conditions.
Across models, the AI label generally yields higher entropy than the Human label, indicating greater decision uncertainty when the same content is disclosed as AI-generated.
Overall, label manipulations modulate model uncertainty.

\begin{figure}[H]
\centering
\includegraphics[width=0.45\textwidth]{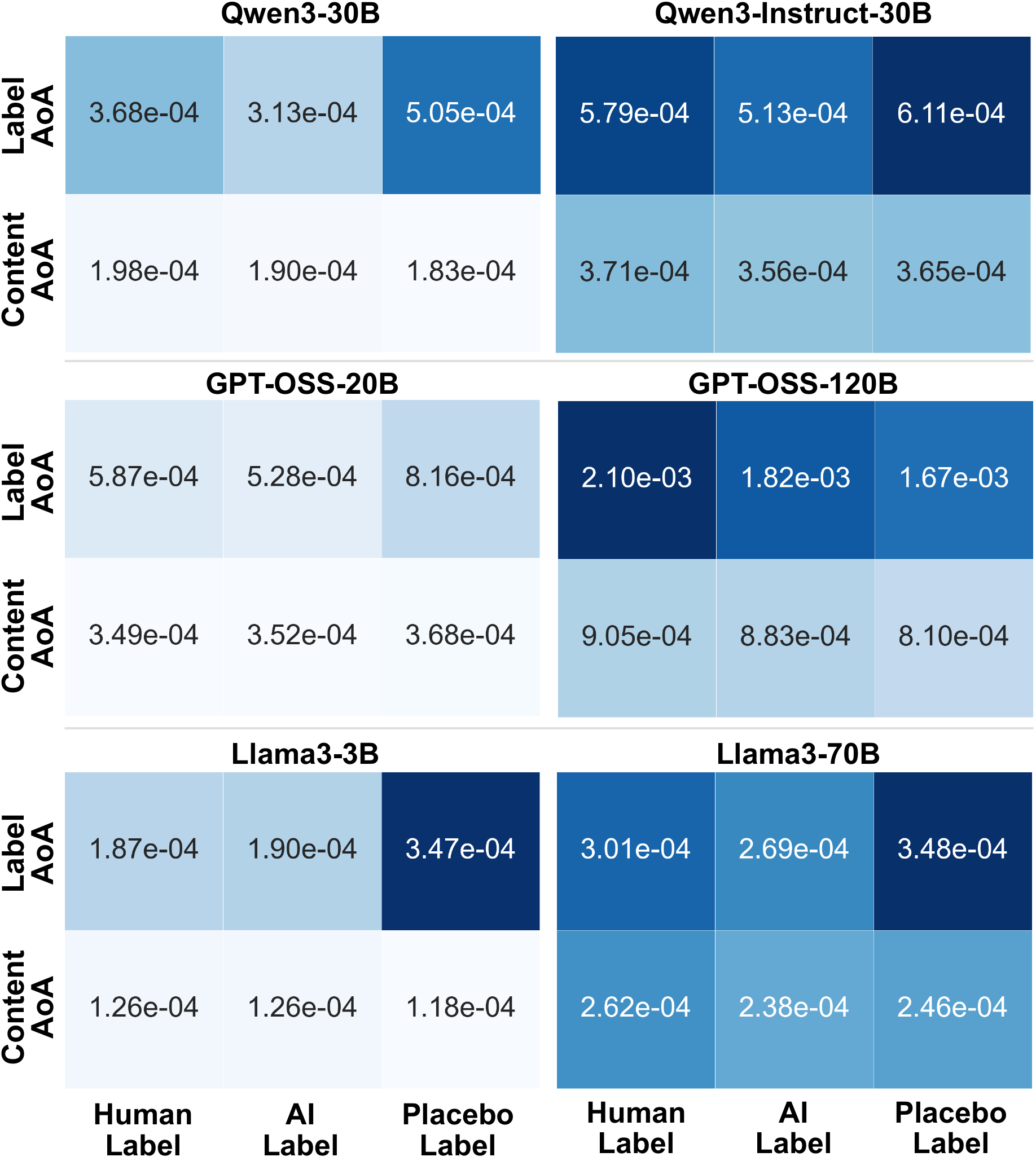}
\caption{
LLM attention distribution density across AOAs (Label AoA vs. Content AoA) and three labels (i.e., Human vs.\ AI vs.\ Placebo) across LLMs.
}
\label{fig:llm_attention_heatmap_3labels}
\vspace{-1.0mm}
\end{figure}

Fig.~\ref{fig:llm_attention_heatmap_3labels} visualizes token-normalized attention density for the Label AoA and Content AoA under three label conditions.
Across models, attention density is higher in the Label AoA than in the Content AoA in all label conditions, consistent with label-dominant processing during judgment.

\section{AI Usage Disclosure}
\label{sec:appendix_ai_disclosure}


We used AI tools for:
(1) language editing (e.g., improving clarity and conciseness) by GPT-5.2.
(2) minor figure polishing (e.g., layout structure and color choice for Fig.~\ref{fig:study_procedure}) by Gemini 3. 
All figures are based on our own data and results. 
All study design, analysis, literature review, and writing were conducted and verified by the authors.


\end{document}